\pdfoutput=1

\documentclass[11pt]{article}

\usepackage[]{emnlp2021}

\usepackage{times}
\usepackage{latexsym}

\usepackage[T1]{fontenc}

\usepackage[utf8]{inputenc}

\usepackage{microtype}

\usepackage{booktabs} 

\usepackage{textcomp}
\usepackage{latexsym}
\usepackage{graphicx} 
\usepackage{microtype}
\usepackage{pifont}
\usepackage[font=small]{caption}
\usepackage[]{subcaption}
\usepackage{verbatim}
\usepackage{framed}
\usepackage{color, colortbl}
\usepackage[colorinlistoftodos]{todonotes}
\usepackage{amssymb}
\usepackage{amsmath}
\usepackage{breqn}
\usepackage{capt-of}
\usepackage{multirow}
\usepackage{algorithmicx}
\usepackage{algorithm}
\usepackage{algcompatible}
\usepackage{paralist}
\usepackage{enumitem}
\usepackage{balance}
\usepackage{tabularx}
\usepackage[all]{nowidow}

\usepackage{listings}

\definecolor{high}{HTML}{FFCE8E}

\title{Improved Multilingual Language Model Pretraining for Social Media Text via Translation Pair Prediction}

\author{
  Shubhanshu Mishra \\
  Twitter, Inc. \\ 
  {\tt smishra@twitter.com} \\ \And
  Aria Haghighi\\
  Twitter, Inc. \\ 
  {\tt ahagighi@twitter.com} \\ 
}

\newcommand{\tpp}{\textsc{TPP}}

\begin{document}
\maketitle
\begin{abstract}
We evaluate a simple approach to improving zero-shot multilingual transfer of mBERT on social media corpus by adding a pretraining task called translation pair prediction (\tpp{}), which predicts whether a pair of cross-lingual texts are a valid translation. Our approach assumes access to translations (exact or approximate) between source-target language pairs, where we fine-tune a model on source language task data and evaluate the model in the target language. In particular, we focus on language pairs where transfer learning is difficult for mBERT: those where source and target languages are different in script, vocabulary, and linguistic typology. We show improvements from TPP pretraining over mBERT alone in zero-shot transfer from English to Hindi, Arabic, and Japanese on two social media tasks: NER (a 37\% average relative improvement in F$_1$ across target languages) and sentiment classification (12\% relative improvement in F$_1$) on social media text, while also benchmarking on a non-social media task of Universal Dependency POS tagging (6.7\% relative improvement in accuracy). Our results are promising given the lack of social media bitext corpus. Our code can be found at: \url{https://github.com/twitter-research/multilingual-alignment-tpp}.

\end{abstract}

\section{Introduction}

\begin{figure}[!htbp]
    \centering
    \includegraphics[width=\linewidth]{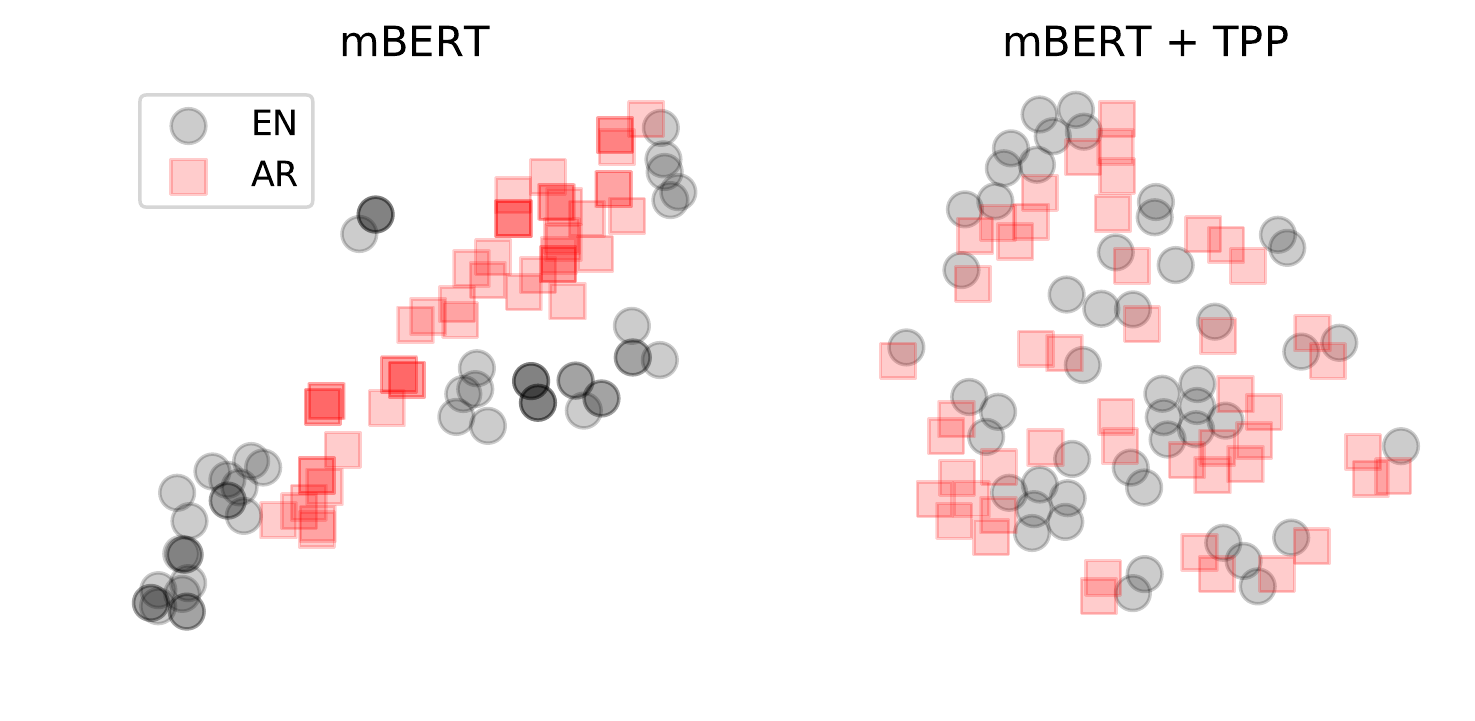}
    \caption{
    \textbf{Comparing Multilingual Representations} Left figure shows mBERT embeddings (T-SNE projection) for parallel sentences from the Tatoeba corpus (Section~\ref{subsec:setup}) in English (EN) and Arabic (AR). These embeddings exhibit distinct language regions, making cross-lingual transfer challenging. The right figure shows the same data after further pretraining using translation pair prediction  (Figure~\ref{fig:model}), which learns representations that are near cross-lingual translations (change in distance showed in Figure \ref{fig:embedding_dist} of appendix). In Section~\ref{sec:eval}, we show this improves cross-lingual transfer performance on several tasks (see Table~\ref{tab:overall-results}). 
    }
    \label{fig:embedding_change}
\end{figure}

Multilingual BERT (mBERT\@; \citealt{devlin-etal-2019-bert}), and other multilingual pretrained language models have been shown to learn surprisingly good cross-lingual representations for a range of NLP tasks \cite{pires2019multilingual,hasoc_2021_journal,mishra-etal-2020-multilingual,Mishra2019HASOC}, despite not having explicit cross-lingual links between their monolingual training corpora \cite{laser, xlnet}. Analysis in \citet{pires2019multilingual} shows that cross-lingual transfer with mBERT works best when transferring representations between source and target languages that share lexical structure (i.e, overlapping word-piece vocabulary) and structural linguistic typology. In some ways, this result is not surprising since mBERT does not see explicit cross-lingual examples and cannot learn how to transform representations to a target language and relies on similar linguistic structure to transfer well. Furthermore, limited work exists on multilingual transfer for social media corpus which tend to be more noisy and shorter compared to traditional NLP corpora. This issue is further exacerbated for transfer between language pairs which use different scripts. 

\begin{figure}
    \centering
    \includegraphics[width=\columnwidth]{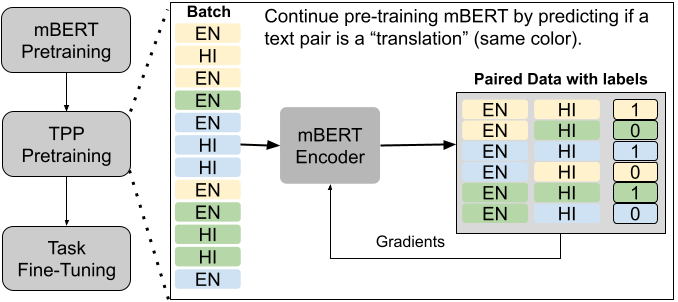}
    \caption{\textbf{Translation Pair Prediction (TPP)} We further pretrain an mBERT model on the TPP task (see Section~\ref{sec:model}). For each batch of aligned source-target text translation pairs, we create a balanced binary dataset consisting of the aligned pairs (positive examples) and a random source-text within the batch (negative examples). We update the mBERT model based on optimizing for binary cross-entropy on this task. In Section~\ref{sec:eval}, we show this added TPP phase improves transfer learning to the target language on downstream tasks. }
    \label{fig:model}
\end{figure}

In this work, we explore whether we can improve multilingual transfer by providing cross-lingual examples and encouraging mBERT to learn aligned representations between source and target texts before task fine-tuning. Concretely, we perform a ``language pretraining" task called \textit{translation pair prediction} (\tpp) where we assume access to a corpus of source-target text pairs and predict whether the pair is a translation of the same content (see Section~\ref{sec:model}). This pretraining phase is intended to occur after standard mBERT pretraining \cite{devlin-etal-2019-bert}, but before task fine-tuning (see Figure~\ref{fig:model}). The intent behind this task is to leverage translation data that exists between a source language with abundant task training data and a target language with little or no task training data in order to improve transfer learning. As with standard mBERT pretraining, one can pretrain mBERT using TPP once and fine-tune for multiple tasks (as long as transferring to the same target language set). The translation pair data doesn't need to be related to the downstream task and we can leverage multiple sources of source-target translations of varying degrees of translation quality (see Section~\ref{subsec:alignment}).

We demonstrate the benefits of adding TPP pretraining in experiments on social media corpus for NER, and sentiment detection, along with a non-social media corpus of universal POS tagging (for comparison) on fine-tuned models from English to Hindi, Japanese, and Arabic (see Section~\ref{sec:eval} for evaluation). We show gains from TPP on all tasks (averaged across target languages) using translation data of varying quality (see Section~\ref{subsec:alignment}).

\paragraph{Related works} \citet{gururangan-etal-2020-dont} argues for the benefits of continued transformer pretraining for better performance on in-domain tasks. Our work closely follows this approach however instead of keeping the Masked Language Modelling (MLM) task on in-domain data, we introduce a new pretraining task (\tpp{}). In \citet{artetxe-schwenk-2019-massively} the authors train an BiLSTM encoder-decoder model on a large parallel corpora of multiple languages and demonstrate zero-shot transfer. In \citet{eisenschlos-etal-2019-multifit} the authors use a domain/task specific fine-tuning to improve zero shot cross lingual performance by inducing labels from the model of \citet{artetxe-schwenk-2019-massively}. Our work is closely related to the work of \citet{wieting-etal-2019-simple,guo-etal-2018-effective} which use a subword averaging and a LSTM encoder. Our approach uses transformers and also focuses evaluation on individual token level task like NER as well as sentence level task. Another closely related work is of \citet{Cao2020Multilingual} which uses word pair alignment with mBERT and evaluates on XNLI task. However, their work does not consider typologically different languages like this work. In \citet{lample2019cross} the authors use Translation Language Modeling task which concatenate the translation pairs in the BERT model and also add a language identifier to each token. \citet{huang-etal-2019-unicoder} include a similar approach as discussed before and get improvement from jointly fine-tuning model with translations of downstream task data. Our approach is comparatively simpler to implement as we don't make any assumption about the encoding model or assumptions about word-level alignments; our approach works independently of the choice of encoder (mBERT here) or the quality of word-alignment models.

\section{Model}
\label{sec:model}
A description of our approach is provided in Figure \ref{fig:model}. The primary exploration in this work is the impact of adding a translation pair prediction (TPP) pretraining phase between traditional mBERT pretraining and downstream task fine-tuning \cite{devlin-etal-2019-bert}. The TPP pretraining task assumes a collection of aligned text pairs $(s, t)$ which are ``translation" pairs between source and target language text pairs $s$ and $t$ respectively. It is assumed that both languages have word-piece coverage in the pretrained mBERT model, but that we primarily have task training data for the source language. After TPP pretraining, we can fine-tune the resulting model for multiple tasks as with standard mBERT for transfer to the target language(s).

The TPP task is a binary classification problem where given an $(s,t)$ pair we predict whether it is a valid ``translation" pair or a random source-target pair using the dot product of mBERT encodings for $s$ and $t$.\footnote{Note that in contrast to some related work \cite{huang-etal-2019-unicoder, lample2019cross} we explore a symmetric task formulation instead of an asymmetric formulation as in the Next Sentence Prediction (NSP) pretraining task where source and target representations are concatenated before prediction.}  More formally, if $f(\cdot)$ represents the encoder mapping a document to the mBERT \texttt{[CLS]} token embedding, we predict if the pair is a translation pair with probability:
\begin{equation*}
    \texttt{is-translation}(s,t) = \sigma \left( f(s)^T f(t) \right)
\end{equation*}
where $\sigma(\cdot)$ is the sigmoid activation function. We create a binary dataset from aligned pairs by sampling a random unaligned $s$ and $t$ pair as a negative example for each aligned pair, creating a balanced dataset. The TPP task is trained using binary cross-entropy against this label and the loss gradient backpropogates through the embedding function $f(\cdot)$ which include mBERT layer parameters. 

After TPP pretraining, we fine-tune on task data in the source language (English in our experiments) and evaluate transfer performance on target languages (Hindi, Japanese, and Arabic in our experiments).

\section{Evaluation}
\label{sec:eval}

\subsection{Experimental setup}
\label{subsec:setup}

In order to evaluate our model we focus on three languages which are dissimilar in vocabulary and syntax from English. 
Furthermore, \textbf{our goal is to improve performance on social media data which is noisier as well as low resource for the above languages and our benchmarked tasks compared to newswire corpus}. Based on these criteria we utilize three languages apart from English which meet the following requirements: a) availability of parallel translation corpora, and b) availability of task data, particularly in the social media domain (NER and sentiment detection here). Based on the above criteria we identify English as our source language and Hindi, Japanese, and Arabic as target languages. We also include POS tagging on Universal Dependencies, which is a non social media benchmark corpus to evaluate our models. More details can be found at: \url{https://github.com/twitter-research/multilingual-alignment-tpp}.

\begin{table}[!htbp]
    \centering
    \begin{tabular}{lrrr}
    \toprule
    Lang pair & Tatoeba & Wikimatrix & Wikidata \\
    \midrule
    en-ar & 28K & 773K & 1.6M  \\
    en-ja & 220K & 480K & 509K \\
    en-hi & 11K & 134K & 77K \\
    \bottomrule
    \end{tabular}
    \caption{Translation pair corpus data sizes used for translation pair prediction pretraining. See Section~\ref{subsec:setup} for details.}
    \label{tab:tpp-data}
\end{table}

\paragraph{Translation pair data} For translation pairs we utilize datasets from the Tatoeba (TT) ~\footnote{\url{https://tatoeba.org} data is under \href{https://creativecommons.org/licenses/by/2.0/fr/}{CC-BY 2.0 FR}.} collection as well as the Wikimatrix (WM) collection \cite{Schwenk2019WikiMatrixM1}. While Tatoeba consists of human generated short sentences, Wikimatrix contains aligned sentences mined using a neural network. We assume that Tatoeba has higher quality across languages compared to Wikimatrix, where many of the translations are either incorrect or not exact translations. We also introduce a new dataset called Wikidata aligned pairs (WD) which can be automatically generated by using pair of language labels and descriptions from Wikidata which are written by human annotators in different scripts. Table~\ref{tab:tpp-data} has details on the dataset size. 

\paragraph{Wikidata aligned pairs (WD)} We collect dataset from Wikidata (\url{https://wikidata.org}). For a given language pair, we consider wikidata items which have wikidata item labels and descriptions in the given pair of languages. We concatenate the label and the description using a space to make a sentence and use the sentences for each language as a translation pair. 

\paragraph{NER data} For our downstream task we use Tweet datasets. For NER we generate a dataset of 100k English Tweets, which is used for training. It consists of Tweets annotated for PERSON, LOCATION, PRODUCT, ORGANIZATION, and OTHER. We also generate test datasets of 2.3k Japanese and 10k Arabic Tweets. All datasets are generated using crowd sourced annotations. For Hindi, we use the Hindi subset of 2008 SSEA shared task \cite{nerhindi}. We report micro avg-F1 score across all classes. We use the standard NER task formulation for BERT style models which use softmax on the first subword for each token to make the prediction \cite{devlin-etal-2019-bert}. The NER models can be further improved by using Conditional Random Fields (CRF) \cite{CRF} layer on the output which often gives good performance even without neural features across all languages \cite{Mishra2020EDNIL}.

\paragraph{Sentiment Detection data} We utilize the SemEval Twitter Sentiment dataset \cite{rosenthal-etal-2017-semeval} with the data split as used in \citet{mishra-2018-metadata-sentiment}. For Japanese, we utilize the 500k Tweet dataset from \citet{suzuki-2019-ja-tweet-sentiment}. For Arabic we use the data from \citet{abdulla-2013-ar-tweet-sentiment}. For Hindi, we use the dataset from SAIL 2015 shared task for Indian languages \cite{patra-2015-hi-sentiment}. We report macro F$_1$ across classes as in \citet{rosenthal-etal-2017-semeval}.

\paragraph{Universal Dependencies POS data} We use the English, Hindi, Japanese, and Arabic subsets from the Universal dependencies data \cite{nivre-etal-2020-universal}. We train using the GUM + PUD subset for English and evaluate on PUD subset for other languages. We do note a shortcoming of our PUD based datasets, i.e. they were created by translating the original text from English, German, French, Italian, or Spanish.\footnote{Details on UD PUD datasets - \url{https://github.com/UniversalDependencies/UD_English-PUD}} This can introduce translation artifacts and can favor translation based pre-training approaches.

\paragraph{Training} All models are trained using the mBERT model available from the HuggingFace library~\cite{huggingface}. For each translation pair data the model is trained for 3 epochs (hyperparameter details in appendix section \ref{sec:training_setup}). For fine-tuning on downstream task we initialize our model with weights from the pretrained models and train it on the English task dataset for 3 epochs. 

\paragraph{TPP using all target languages} We also experiment with a TPP model which was pretrained using all the target language pair data from Tatoeba which were equally sampled. We denote this setting by \textsc{ALL} in Table~\ref{tab:overall-results} for our results. Even though this model may not perform as well as pretraining with TPP for a single target language at a time, it allows us to reduce pretraining and inference time, as well as reduce the number of models to manage. 

\begingroup
\setlength{\tabcolsep}{3.5pt} 
\begin{table}[]
    \centering
    \begin{tabular}{lrr|rr|rr}
    \toprule
    {} & \multicolumn{2}{c}{Hindi} & \multicolumn{2}{c}{Japanese} & \multicolumn{2}{c}{Arabic} \\
    \midrule
    \textbf{NER} &    F$_1$ & $\Delta$\% &    F$_1$ & $\Delta$\% &    F$_1$ & $\Delta$\% \\
    \midrule
    mBERT               &  21.1 &   0.0 &  16.5 &   0.0 &  32.1 &   0.0 \\
    +TPP (\textsc{One})      & \textbf{24.3} &  15.2 &  \textbf{29.9} &  81.4 &  \textbf{39.4} &  22.8 \\
    +TPP (\textsc{ALL}) &  23.2 &  10.3 &  27.4 &  66.4 &  38.5 &  19.9 \\
    \toprule
    \textbf{Sentiment} &    F$_1$ & $\Delta$\% &    F$_1$ & $\Delta$\% &    F$_1$ & $\Delta$\% \\
    \midrule
    mBERT               &  31.7 &   0.0 &  55.0 &   0.0 &  51.5 &   0.0 \\
    +TPP (\textsc{One})       & \textbf{32.7} &   3.0 & 66.4 &  20.6 & 58.3 &  13.2 \\
    +TPP (\textsc{All}) &  32.4 &   2.3 &  \textbf{67.7} &  23.1 &  \textbf{58.5} &  13.7 \\
    \toprule
    \textbf{UD POS} &    acc. & $\Delta$\% &    acc. & $\Delta$\% &    acc. & $\Delta$\% \\
    \midrule
    mBERT               &   67.4 &   0.0 &  52.7 &  0.0 &  64.0 &   0.0 \\
    +TPP (\textsc{One}) &   \textbf{71.5} &   6.0 & \textbf{57.6} &   9.2 & \textbf{67.1} &   4.8 \\ 
    +TPP (\textsc{All}) &  66.4 &  -1.5 &  52.7 &  0.1 &  65.0 &   1.5 \\
    \bottomrule
    \end{tabular}

    \caption{Results on zero-shot transfer learning for NER (F$_1$), sentiment detection (F$_1$), and UD POS tagging (Accuracy) from English to Hindi, Japanese, and Arabic. We compare fine-tuning standard mBERT to fine-tuning mBERT further pretrained on the TPP task (Section~\ref{sec:model}). The \textsc{One} variant pretrains a separate model for each target language; the \textsc{All} variant is a single pretraiend model for all target languages. We use $\Delta\%$ to display relative improvement.
    }
    \label{tab:overall-results}
\end{table}
\endgroup

\begin{figure}[!tb]
    \centering
    \includegraphics[width=0.8\linewidth]{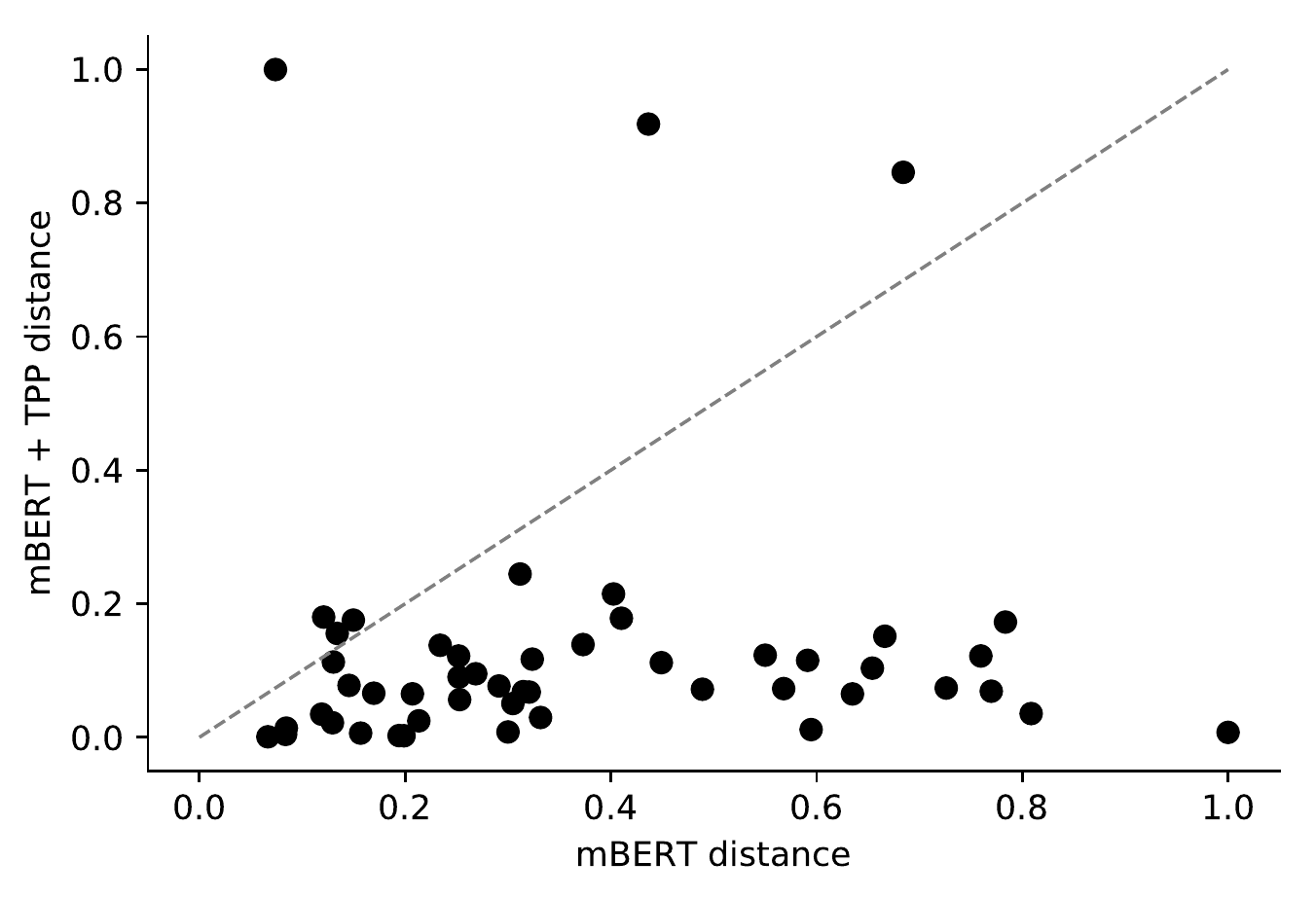}
    \caption{Change in embedding distance of EN-AR translation pairs shown in Figure \ref{fig:embedding_change}}
    \label{fig:embedding_dist}
\end{figure}

\paragraph{Change in embedding distance after alignment} In Figure \ref{fig:embedding_dist} we show how the normalized (by max) embedding distances change beetween sentence pairs shown in Figure \ref{fig:embedding_change} after applying \tpp{}.We find that alignment pairs are closer than their distance in mBERT. 

\subsection{Zero-shot improvement via TPP}
As can be seen in Table~\ref{tab:overall-results}, adding \tpp{} pretraining improves performance for both NER and sentiment detection across all languages. The universal POS gains are consistent, but more modest; we hypothesize this is because POS-tagging reilies more heavily on the language-neutral components of mBERT representations~\cite{mbertPOS}. The greatest improvement of our models is for Japanese NER for which the model improves by 81\% and an absolute improvement of around 15\% in $F_1$ score. We hypothesize that NER benefits the most from some our \tpp{} sources (Wikimatrix and Wikidata) which contain many paired entity descriptions. Furthermore, the Tatoeba dataset comprises of short sentence which are not likely to contain sentiment information, especially sentiment bearing content likely to be shared on social media platforms.

\subsection{Improvement using combinations of alignment pairs and quality of alignments}
\label{subsec:alignment}
In this section we discuss the impact of translation datasets available for our languages as well as how combining them for the \tpp~task leads to improved performance. We focus on NER $F_1$ and relative improvements (denoted $\Delta\%$), but other tasks also see similar variation for each language (see Table \ref{tab:translation-quality-results} in appendix). We find that increasing the alignment pair data or using higher quality alignment data yield improvements. Using single-corpus translation pairs alone leads to significant improvements for all languages ($\Delta \% $ HI=9.6, JA=68.6, AR=19.3). Tatoeba results in most prominent improvements (HI=9.6, JA=68.6), while Wikimatrix has the best performance for AR (19.3). We assume this is related to larger size of the data, e.g. JA has 10 times more data in Tatoeba as compared to AR and HI. 

\begingroup
\setlength{\tabcolsep}{3.5pt} 
\begin{table}[!tbp]
    \centering
    \begin{tabular}{lrr|rr|rr}
    \toprule
    {} & \multicolumn{2}{c}{Hindi} & \multicolumn{2}{c}{Japanese} & \multicolumn{2}{c}{Arabic} \\
    \midrule
    \textbf{NER} &    $F_1$ & $\Delta$\% &    $F_1$ & $\Delta$\% &    $F_1$ & $\Delta$\% \\
    \midrule
    mBERT     &  21.1 &   0.0 &  16.5 &   0.0 &  32.1 &   0.0 \\
    \midrule
    +TPP (\textsc{TT})        &  23.1 &   9.6 &  27.8 &  68.6 &  36.3 &  13.2 \\
    +TPP (\textsc{WD})        &  22.4 &   6.3 &  26.5 &  60.8 &  36.9 &  15.0 \\
    +TPP (\textsc{WM})        &  21.6 &   2.6 &  27.7 &  68.3 &  38.3 &  19.3 \\
    \midrule
    +TPP (\textsc{BP}) &  24.3 &  15.2 &  29.9 &  81.4 &  39.4 &  22.8 \\
    \midrule
    +TPP (\textsc{ALL}) &  23.2 &  10.3 &  27.4 &  66.4 &  38.5 &  19.9 \\
    \bottomrule
    \toprule
    \textbf{Sentiment} &   $F_1$ & $\Delta$\% &   $F_1$ & $\Delta$\% &   $F_1$ & $\Delta$\% \\
    \midrule
    mBERT     & 31.7 &   0.0 & 55.0 &   0.0 & 51.5 &   0.0 \\
    \midrule
    +TPP (\textsc{TT})        & 31.8 &   0.3 & 62.4 &  13.5 & 58.3 &  13.2 \\
    +TPP (\textsc{WD})        & 30.8 &  -2.9 & 50.2 &  -8.7 & 53.0 &   3.0 \\
    +TPP (\textsc{WM})        & 32.7 &   3.0 & 63.2 &  14.8 & 54.7 &   6.4 \\
    \midrule
    +TPP (\textsc{BP})      & 32.0 &   0.9 & 66.4 &  20.6 & 55.3 &   7.5 \\
    \midrule
    +TPP (\textsc{ALL}) & 32.4 &   2.3 & 67.7 &  23.1 & 58.5 &  13.7 \\
    \bottomrule
    \toprule
    \textbf{UD POS} &   acc. & $\Delta$\% &   acc. & $\Delta$\% &   acc. & $\Delta$\% \\
    \midrule
    mBERT     & 67.4 &   0.0 & 52.7 &   0.0 & 64.0 &   0.0 \\
    \midrule
    +TPP (\textsc{TT})        & 65.1 &  -3.5 & 54.0 &   2.4 & 66.7 &   4.1 \\
    +TPP (\textsc{WD})        & 70.5 &   4.5 & 53.0 &   0.5 & 66.4 &   3.7 \\
    +TPP (\textsc{WM})        & 70.4 &   4.3 & 54.4 &   3.1 & 65.4 &   2.2 \\
    \midrule
    +TPP (\textsc{BP})      & 71.5 &   6.0 & 57.6 &   9.2 & 67.1 &   4.8 \\
    \midrule
    +TPP (\textsc{ALL}) & 66.4 &  -1.5 & 52.7 &   0.1 & 65.0 &   1.5 \\
    \bottomrule
    \end{tabular}
    \caption{Impact of translation quality and using combination of translations in NER, Sentiment, and UD POS examples task performance after TPP (F$_1$ = micro F$_1$ score for NER, macro F$_1$ for sentiment, acc. = accuracy for UD POS). $\Delta \%$ is the $\%$ change in evaluation score compared to mBERT. Translation pairs used for each \tpp{} task are shown in parenthesis, TT=Tatoeba, WD=Wikidata, WM=WikiMatrix, BP=best pairing of TT, WD, WM, and ALL=TT pairs from all languages, equally sampled.}
    \label{tab:translation-quality-results}
\end{table}
\endgroup

Next, we move to sequencing pairs of \tpp{} tasks from each translation dataset for a given language pair. We find that as we pair the \tpp datasets the performance of the models continues to improve (Table \ref{tab:translation-quality-results}) but it depends on the previous checkpoint we start from as well as the quality of the new \tpp{} data. Our experiments revealed that using the best performing single pairs in their order of performance individually improves performance significantly ($\Delta \% $ HI=15.2, JA=81.4, AR=22.8). For Sentiment and UD POS we observe similar gains, except in few case there is slight degradation. 

\paragraph{Note on translation quality} In terms of translation quality, Tatoeba is likely to be the most accurate as it is manually curated. Next, Wikidata is likely to be higher quality for HI compared to Wikimatrix as Wikimatrix is auto generated using a model and hence likely to perform worse on low resource languages. For AR and JA we can expect Wikimatrix to be higher quality as these languages have larger Wikipedia size and hence likely to have better quality representation in pretrained LMs. 

\section{Conclusion}
We evaluated the use of translation pair prediction (TPP) as a pretraining task which can improve the performance of multilingual language models for cross-lingual zero-shot transfer from English. This pretraining task is performed after standard mBERT pretraining and the resulting model can be task fine-tuned for cross-lingual transfer similar to mBERT. We show significant improvement in zero-shot transfer by adding TPP pretraining on NER and sentiment detection on social media tasks for three languages (Hindi, Japanese, and Arabic) which are dissimilar in vocabulary and syntax from English.

\bibliographystyle{acl_natbib}
\bibliography{anthology,custom}

\appendix
\clearpage

\section{Experiment}
\label{sec:experiment_details}

\paragraph{Code availability} Our  model training code and evaluation code will be released at: \url{https://github.com/twitter-research/multilingual-alignment-tpp}.

\paragraph{Training setup}
\label{sec:training_setup} Our default parameters are as follows: We use Adam optimizer with 500 warmup steps=500 and weight decay of 0.01. Our batch size is 16. Models were trained using 2 NVIDIA v100 GPUs. Training time for \tpp{} ranged from 30 mins to 32hrs for different settings. NER training time was around 1.5 hrs. Sentiment training time was 1 minute, and UD POS training time was also 1 minute.

\paragraph{Evaluation}
\label{sec:evaluation_setup} For NER tasks we used micro-averaged F1 score  as implemented in the seqeval \footnote{\url{https://github.com/chakki-works/seqeval}} library. For sentiment tasks we use document level micro-averaged F1 score implemented in scikit-learn library \footnote{\url{https://scikit-learn.org/}}. For UD-POS tasks we used token level accuracy as implemented in scikit-learn library.

\section{Data}
\label{sec:data_details}
\subsection{Data sources}

\paragraph{NER}
\begin{enumerate}
    \item SSEA Hindi - From the Workshop on NER for South and South East Asian Languages
\url{http://ltrc.iiit.ac.in/ner-ssea-08/index.cgi?topic=5}
    \item Tweets annotated for in English, Japanese, and Arabic with entities from the following types: PERSON, LOCATION, PRODUCT, ORGANIZATION, and OTHER.
\end{enumerate}

\paragraph{Sentiment} For all sentiment datasets we only consider the positive and negative examples and exclude the neutral class if present. 
\begin{enumerate}
    \item Japanese - Available at: \url{http://www.db.info.gifu-u.ac.jp/data/Data_5d832973308d57446583ed9f}
    \item Arabic - Available at: \url{https://archive.ics.uci.edu/ml/datasets/Twitter+Data+set+for+Arabic+Sentiment+Analysis}
    \item Hindi - Available at: \url{http://amitavadas.com/SAIL/data.html}
\end{enumerate}
\end{document}